\documentclass[sigconf,screen]{acmart}
\AtBeginDocument{%
  }

\setcopyright{acmlicensed}
\copyrightyear{2025}
\acmYear{2025}
\acmDOI{10.1145/3746027.3754794}
\acmConference[MM '25] {Proceedings of the 33rd ACM International Conference on Multimedia}{October 27--31, 2025}{Dublin, Ireland.}
\acmBooktitle{Proceedings of the 33rd ACM International Conference on Multimedia (MM '25), October 27--31, 2025, Dublin, Ireland}
\acmISBN{979-8-4007-2035-2/2025/10}

\usepackage{multirow}

\usepackage{amssymb} 

\begin{document}

\title{Towards Measuring and Modeling Geometric Structures in Time Series Forecasting via Image Modality}


\author{Mingyang Yu}
\authornote{Equal contribution.}
\affiliation{
  \institution{East China Normal University}
  \city{Shanghai}
  \country{China}}
\email{yumingyang@stu.ecnu.edu.cn}

\author{Xiahui Guo}
\authornotemark[1]
\affiliation{
  \institution{East China Normal University}
  \city{Shanghai}
  \country{China}}
\email{xhguo@stu.ecnu.edu.cn}

\author{Peng Chen}
\affiliation{
  \institution{East China Normal University}
  \city{Shanghai}
  \country{China}}
\email{pchen@stu.ecnu.edu.cn}

\author{Zhenkai Li}
\affiliation{%
  \institution{East China Normal University}
  \city{Shanghai}
  \country{China}}
\email{10225101541@stu.ecnu.edu.cn}

\author{Yang Shu}
\authornote{Corresponding author.}
\affiliation{
  \institution{East China Normal University}
  \city{Shanghai}
  \country{China}}
\email{shuyang5656@gmail.com}
\begin{abstract}
  Time Series forecasting is critical in diverse domains such as weather forecasting, financial investment, and traffic management. While traditional numerical metrics like mean squared error (MSE) can quantify point-wise accuracy, they fail to evaluate the geometric structure of time series data, which is essential to understand temporal dynamics. To address this issue, we propose the time series Geometric Structure Index (TGSI), a novel evaluation metric that transforms time series into images to leverage their inherent two-dimensional geometric representations. However, since the image transformation process is non-differentiable, TGSI cannot be directly integrated as a training loss. We further introduce the Shape-Aware Temporal Loss (SATL), a multi-component loss function operating in the time series modality to bridge this gap and enhance structure modeling during training. SATL combines three components: a first-order difference loss that measures structural consistency through the MSE between first-order differences, a frequency domain loss that captures essential periodic patterns using the Fast Fourier Transform while minimizing noise, and a perceptual feature loss that measures geometric structure difference in time-series by aligning temporal features with geometric structure features through a pre-trained temporal feature extractor and time-series image autoencoder. Experiments across multiple datasets demonstrate that models trained with SATL achieve superior performance in both MSE and the proposed TGSI metrics compared to baseline methods, without additional computational cost during inference.
\end{abstract}

\ccsdesc[500]{Computing methodologies~Temporal reasoning}

\keywords{Time Series Forecasting, Evaluation Metric, Loss Function}


\maketitle

\section{Introduction}
\begin{figure}[h]  
    \centering      
    \includegraphics[width=0.47\textwidth]{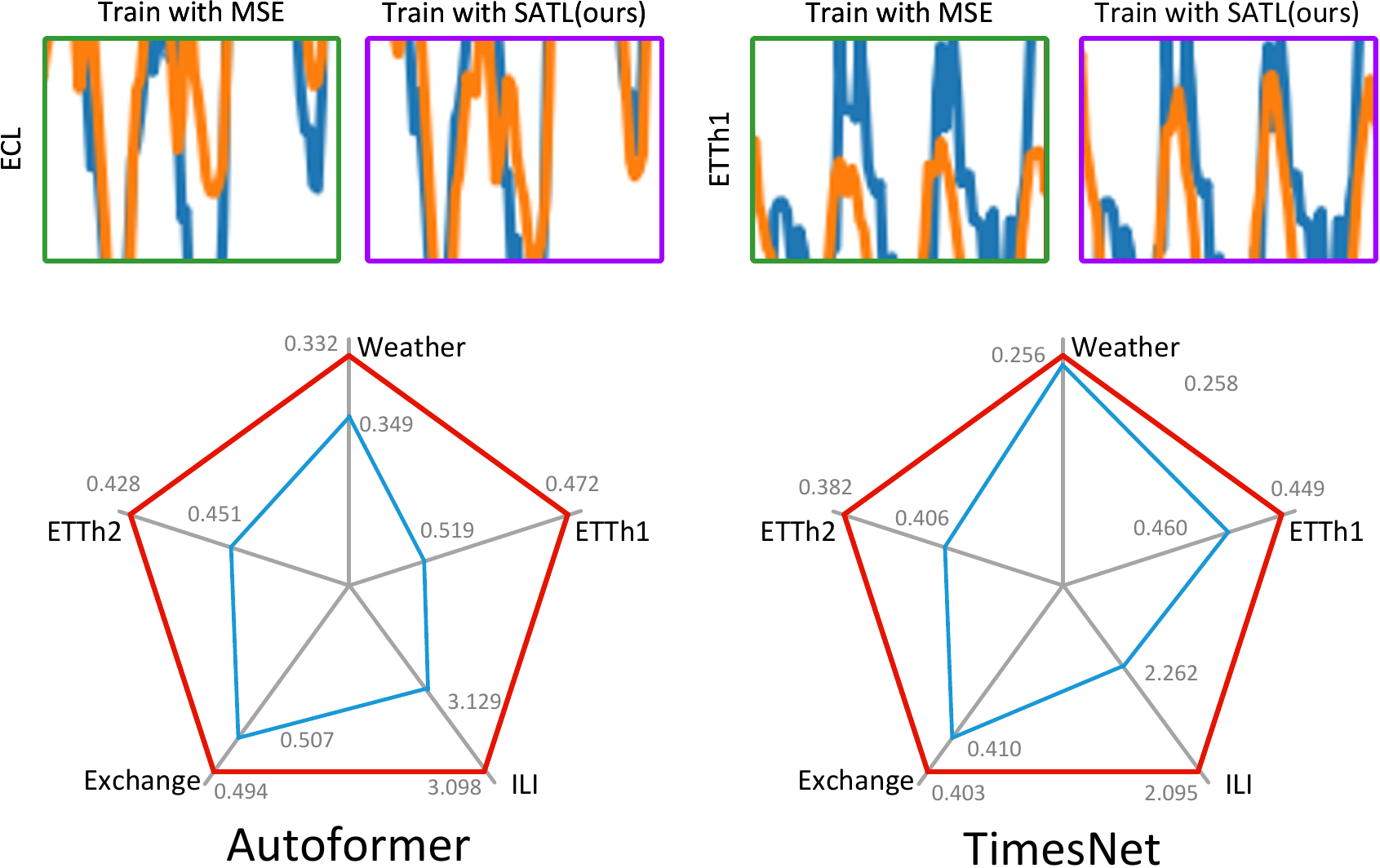}  
    \caption{Time series forecasting comparison of Autoformer and TimesNet. Top row displays prediction visualizations (MSE-trained vs. SATL-trained models) with ground truth (blue) and forecasts (yellow), where closer alignment indicates higher accuracy. Bottom row shows quantitative MSE comparisons. SATL achieves superior performance without additional inference cost.}  
    \label{fig:1}  
    \vspace{-5mm}
\end{figure}
With the rapid development of Multimedia and the Internet of Things, a massive amount of time series data is being generated, driving significant progress in the time series forecasting task. Time series forecasting aims to predict future information based on historical information, thereby supporting decision-making and demonstrating broad application prospects ~\cite{wu2023interpretable, granger2014forecasting, yin2021deep}. Due to the dynamic variability of time series, modeling these time series data can be quite challenging. The two key aspects of time series data are its shape and values \cite{keogh2003efficiently,esling2012time}. Although the numerical aspect has been the main focus for evaluation, the shape of the data is often neglected, even though it is crucial for comprehending how the data changes over time.

\begin{figure}[t]
\centering 
\includegraphics[width=0.47\textwidth]{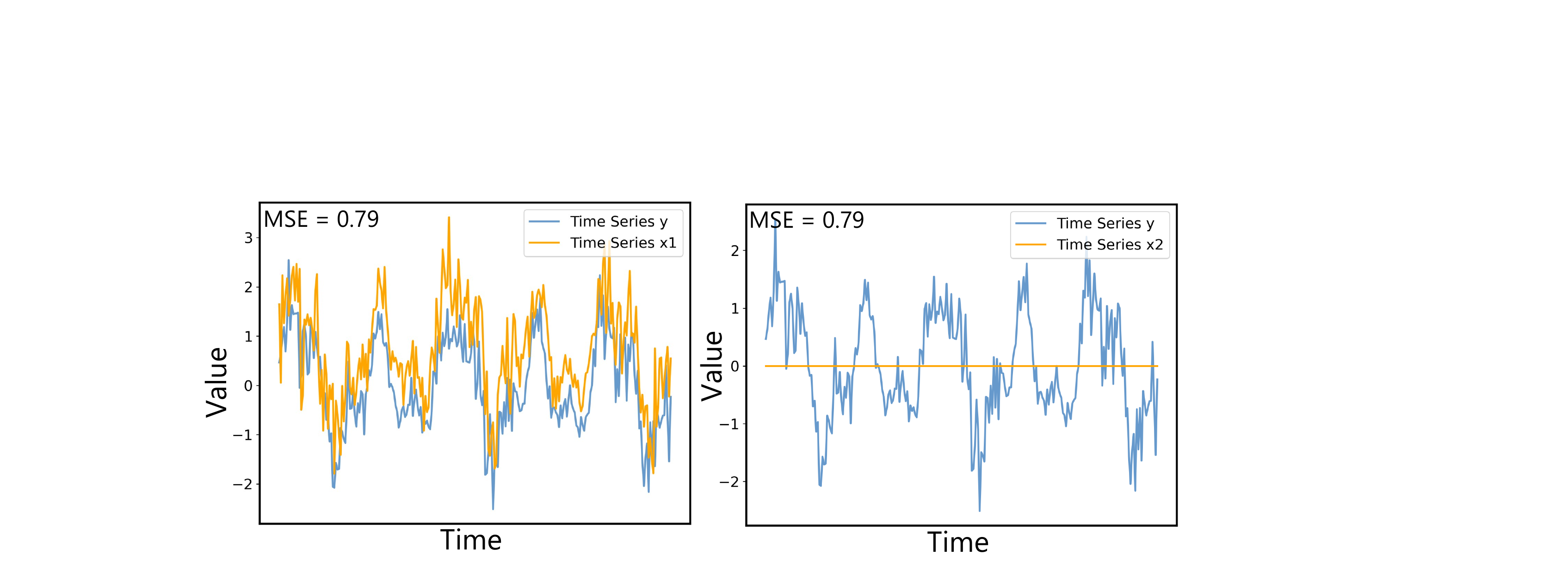} 
    \caption{Comparison of time series y with $x_1$ (left) and $x_2$ (right). Both have MSE = 0.79, but $x_1$ shows a closer geometric resemblance to y.}  
    \label{fig:2}  
    \vspace{-6mm}
\end{figure}

In current time series forecasting tasks, metrics like Mean Squared Error (MSE) and Mean Absolute Error (MAE) are commonly used~\cite{lim2021time,jadon2024comprehensive}. However, these metrics fail to capture the shape of time series data, which reflects its underlying geometric structure. As illustrated in Figure~\ref{fig:2}, we generate a time series $y$ that exhibits fundamental periodicity along with added noise. Two additional time series are created, where $x_1$ shares the same period as $y$ but is subject to a vertical shift and different noise levels. In contrast, $x_2$ is a time series consisting of zeros. When we calculate the MSE between the pairs ($y$, $x_1$) and ($y$, $x_2$), we find that both pairs yield an MSE of 0.79. However, the geometric structure of $x_1$ is significantly more similar to that of $y$ than that of $x_2$. This observation highlights a critical limitation of MSE as a numerical evaluation metric as it fails to adequately capture the geometric similarity between two time series even if MSE values are identical. Similarly, other loss functions based on the $L_p$ norm, such as MAE, also suffer from the same weakness as MSE. Some alternative methods, such as Dynamic Time Warping (DTW~\cite{rakthanmanon2012searching}), have tried to consider the geometric structure of time series data, but they still rely on the temporal dimension, making it difficult to fully capture geometric information~\cite{cuturi2017soft,mensch2018differentiable,abid2018learning}.

To solve this problem, we propose evaluating the geometric structure of time series data by transforming it into images. We introduce the Temporal Geometric Structure Index (TGSI), a novel metric specifically designed for visualizing and assessing the geometric structure of time series data. We transform time series into images, where the horizontal axis represents time and the vertical axis represents a range of possible values. The TGSI measures structural similarity between these corresponding images of time series, focusing on luminance and covariance components. This approach allows for a more detailed analysis of temporal patterns, effectively capturing geometric structural information that traditional metrics, such as MSE, may overlook.

Besides the metrics focusing on geometric structures for performance evaluation, we further explore loss functions to enhance the structure modeling of models during training. In time series forecasting tasks, the commonly used loss function is MSE. As mentioned earlier, this type of loss function fails to capture the geometric structure of time series, leading to predictions that overlook this aspect~\cite{lee2022tilde}. However, the TGSI we propose, which operates in the image modality, cannot be directly used as a loss function for training. This is because the process of transforming time series into image modality is non-differentiable and thus cannot be backpropagated~\cite{rumelhart1986learning}.

To address this, we design a set of loss function called Shape-Aware Temporal Loss (SATL), which reflects geometric structure similarity using only the time series modality. SATL consists of three components: (1) first-order difference loss, which measures the MSE between the first-order differences of the predicted and ground truth time series to emphasize structural consistency. (2) Frequency domain loss, which uses the Fast Fourier Transform to capture essential periodic patterns while minimizing noise. (3) Perceptual feature loss, which specifically measures differences in geometric structure. We train an autoencoder for time series images to extract image features. Subsequently, a temporal feature extractor is trained to extract temporal features, which are then aligned with the image features. This alignment enables the temporal feature extractor to also capture image features. After training the temporal feature extractor, we employ it in the forecasting task to extract features from both ground truth and predicted sequences, then compute the perceptual loss based on their feature differences.

In conclusion, our work makes the following key contributions:
\begin{itemize}
    \item We design TGSI, a specialized evaluation metric for assessing the geometric structure similarity of time series images. This metric complements and enhances the evaluation framework for time series prediction tasks by addressing previously overlooked aspects.
    \item We propose SATL, a loss function that operates in the time series modality but is capable of capturing geometric structure differences of time series. This plug-and-play design enables seamless integration with existing models without architectural modifications.
    \item Our experiments compare models trained with SATL against conventional MSE training across multiple datasets. As shown in Figure~\ref{fig:1}, SATL improves prediction quality, achieving better performance on both MSE and TGSI metrics. These improvements come without additional computational cost during inference.
\end{itemize}

\section{Related Work}
\subsection{Time Series Forecasting}
Time Series Forecasting aims to predict future values by capturing the relationships between past and future data. Statistical models are primitive methods and they are mainly based on exponential smoothing and its variants~\cite{hyndman2008automatic}. Deep learning models became the mainstream and among those recurrent neural networks (RNNs) model the temporal dependency~\cite{chung2014empirical,wen2017multi,cirstea2019graph,kieu2022anomaly}. TimesNet~\cite{wu2022timesnet} transforms one-dimensional time series into a two-dimensional space and captures multi-period features through convolution.

In recent years, there are plenty of works that tried to apply Transformer models to forecast long-term time series and they have demonstrated exceptional capabilities due to capturing the long-term temporal dependencies adaptively with the attention mechanism~\cite{chen2024pathformer,wang2025towards,wang2025lightgts}. Autoformer~\cite{wu2021autoformer} proposed the ideas of decomposition and auto-correlation. PatchTST~\cite{nie2022time} employs patching and channel independence, showing that the Transformer architecture has its potential with proper adaptation. Furthermore, models based on multiple-layer projection (MLP) also show effective performance~\cite{oreshkin2019n,challu2023nhits,wang2024timemixer,wang2024timemixer}. 

Recent studies have explored converting time series into image representations and leveraging computer vision techniques to extract visual structural features from time series for enhanced forecasting performance~\cite{hatami2018classification,li2020forecasting,yang2024vitime,xu2025can,chen2024visionts}.

\subsection{Loss Functions and Metrics}

Currently, Mean Squared Error(MSE), Mean Absolute Error(MAE) are mainstream Loss Functions used to train forecasting models and they are classical Metrics for evaluating. As Loss Functions, they only possess numerical capability, lacking the ability to guide models in learning geometric structural patterns of time series data. As Metrics, they disregard the relationships between different points, failing to capture shape-related information~\cite{le2019shape}.

Recently, a novel Loss Function called TILDE-Q~\cite{lee2022tilde} was proposed, which considers not only amplitude and phase changes but also allows models to capture the shape of time series. However, it fails to model geometric structural information which can't be extracted from the time series modality itself from the numerical perspective. The structural similarity index measure (SSIM~\cite{wang2004image}) is an outstanding method to measure the similarity between two images by assessing luminance, contrast. time series sequence is one-dimensional. The application of metrics from the image domain to the time series domain remains long unexplored, but we fill this gap by applying the idea from SSIM.

\section{Temporal Geometric Structural Index}
The Structural Similarity Index Measure (SSIM~\cite{wang2004image}) effectively captures structural information in natural images, but its direct application to time series images is limited due to their distinct distribution characteristics. To address this limitation, we propose the Temporal Geometric Structural Index (TGSI) as a novel metric specifically tailored for evaluating the geometric structural similarity of time series data. TGSI is specifically designed for time series data visualization and structure assessment. In the following, we first explain the time-to-image transformation and then formally define the TGSI metric, accompanied by an analysis of its design choices and practical advantages.

\subsection{Transformation of time series to Images}
\begin{figure*}[t]  
    \centering      
    \includegraphics[width=0.99\textwidth]{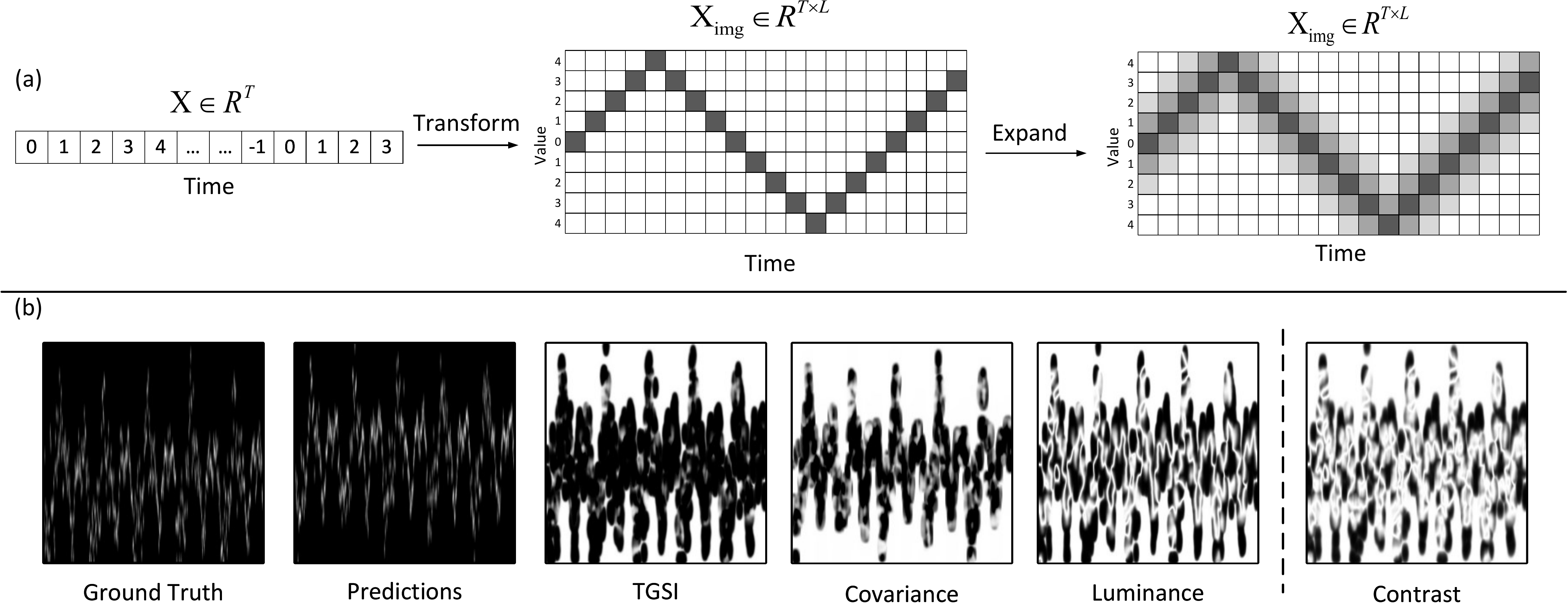}  
    \caption{(a) time series to image transformation. (b) Left of dashed line: Visualization of conversion results with TGSI and its components; Right of dashed line: Visualization of the contrast part for this example.}  
    \label{fig:3}  
\end{figure*}

As illustrated in Figure~\ref{fig:3}(a), time series data is transformed into an image where the horizontal axis represents time, and the vertical axis represents a range of possible values of each time step. The transformation process begins with normalization to ensure consistent intensity scaling across sequences. For each time point, the corresponding value determines the vertical position of the activated pixel in the image plane, establishing a direct mapping between temporal amplitude and spatial position. Initially, the temporal sequence corresponds to a single line of activated pixels in the image. To enhance the structural representation, this line is vertically expanded, creating a gradient-like effect where the pixel brightness decreases as the distance from the original line increases. This expansion serves two critical purposes: 1) it improves robustness by allowing tolerance to small temporal misalignments and value variations, and 2) it prevents structural information loss in covariance computation that would occur with sparse single-line representations. The expanded image effectively encodes the temporal data as a probability-like distribution, where brighter pixels indicate a higher likelihood of the temporal value at that point. 

\subsection{Definition and Analysis of TGSI}
The TGSI metric quantifies the structural similarity between two time series images, with values ranging from -1 to 1, where higher values indicate greater similarity. It is defined as the product of two components: luminance and covariance. Unlike SSIM, the contrast component is not considered for the following reasons: First, the vertical intensity decay during the time-to-image transformation enforces a fixed variance along the y-axis, rendering the contrast component redundant in this dimension. Second, in the temporal dimension, the variance exhibits strong coupling with the luminance component. Furthermore, we visualize the luminance and contrast components in Figure~\ref{fig:3}(b), revealing consistent patterns that substantiate this relationship. The TGSI metric is formally defined as:
\begin{equation}
\mathrm{TGSI}(x, y) = l(x, y) \cdot s(x, y),
\end{equation}
where $l(x, y)$ represents the luminance component, which reflects the probability of occurrence of the temporal values by capturing the brightness consistency across corresponding regions of the time series images, and $s(x, y)$ represents the covariance component, which assesses the structural relationships by evaluating how variations in the pixel intensities correlate between the two images.

\paragraph{Luminance Component:} The luminance component captures the brightness consistency between two time series images $x_{img}$ and $y_{img}$, and is defined as:
\begin{equation}
l(x, y) = \frac{2\mu_x \mu_y + C_1}{\mu_x^2 + \mu_y^2 + C_1},
\end{equation}
where $\mu_x$ and $\mu_y$ are the mean pixel intensities for $x_{img}$ and $y_{img}$, respectively. The constant $C_1$ is introduced to avoid division by zero, defined as $C_1 = (K_1 L)^2$, with $K_1 = 0.01$ and $L$ being the dynamic range of pixel values. The luminance component reflects the probability-like distribution of temporal data encoded in the brightness of the transformed image.

\paragraph{Covariance Component:} The covariance component assesses the structural similarity between $x_{img}$ and $y_{img}$, and is defined as:
\begin{equation}
s(x, y) = \frac{\sigma_{xy} + C_2}{\sigma_x \sigma_y + C_2},
\end{equation}
where $\sigma_{xy}$ is the covariance between $x_{img}$ and $y_{img}$, and $\sigma_x$ and $\sigma_y$ are the standard deviations of $x_{img}$ and $y_{img}$, respectively. The constant $C_2$ is defined as $C_2 = \frac{(K_2 L)^2}{2}$, where $K_2 = 0.03$.

To ensure that the covariance component accurately captures the structural relationships, the image is downscaled by a factor before calculating $\sigma_{xy}$, $\sigma_x$, and $\sigma_y$. This step compensates for the vertical expansion applied during the time-to-image transformation, enabling the covariance term to focus on the geometric structures rather than the intensity gradient.

To illustrate the effectiveness of TGSI, consider the example shown in Figure~\ref{fig:2}. Calculating the TGSI values yields $\mathrm{TGSI}(y, x_1) = 0.5212$ and $\mathrm{TGSI}(y, x_2) = 0.2080$. These results demonstrate that $x_1$ has a stronger structural similarity to $y$ than $x_2$, showcasing TGSI's capability to effectively capture geometric structural differences in temporal sequence images, whereas traditional metrics like MSE fail to distinguish their geometric structural similarities.

\section{Shape-Aware Temporal Loss}
\label{sec_satl}

To effectively capture the geometric structural information of time series modalities during training, it is essential to use an appropriate loss function. However, since we assess geometric structural differences using image modalities in TGSI, and the process of transforming time series modalities into image modalities cannot support backpropagation\cite{rumelhart1986learning}, we must rely on the loss function directly applied to time series modalities. Consequently, we design the Shape-Aware Temporal Loss (SATL), which achieves this goal by leveraging three complementary components. These components work together to ensure that the loss function is sensitive to temporal patterns, frequency-domain information, and perceptual features in time series data.

\subsection{First-Order Temporal Loss}
The first component of the loss measures the MSE between the first-order differences of the predicted and ground truth time series. The first-order difference of a time series captures the changes between consecutive time points, which are critical for understanding the shape and structure of the sequence. 

This approach is motivated by the observation that time series with identical shapes but shifted along the vertical axis should still be considered geometrically similar. The first-order difference reflects shape consistency by emphasizing local trends and variations in the sequence, making it less sensitive to absolute values. By comparing the first-order differences, the loss function inherently focuses on the structural and geometric features of the time series.

Let $\mathbf{x} \in \mathbb{R}^{T \times N}$ and $\mathbf{y} \in \mathbb{R}^{T \times N}$ represent the predicted and ground truth time series, respectively, where $N$ is the number of variables in the time series and $T$ is the length of the sequence. The first-order differences of the sequences are defined as:
\begin{equation}
\mathbf{diff}_x(t) = \mathbf{x}(t+1) - \mathbf{x}(t),  \quad t = 1, 2, \dots, T-1.
\end{equation}

The first-order temporal difference loss is then expressed as:
\begin{equation}
\mathcal{L}_\text{diff} = \frac{1}{(T-1)} \sum_{t=1}^{T-1}  \left( \mathbf{diff}_x(t) - \mathbf{diff}_y(t) \right)^2.
\end{equation}

This component emphasizes variations in the time series data, allowing the model to prioritize the consistency of shape and structure while effectively capturing temporal dynamics.

\begin{figure*}[h]  
    \centering      
    \includegraphics[width=0.99\textwidth]{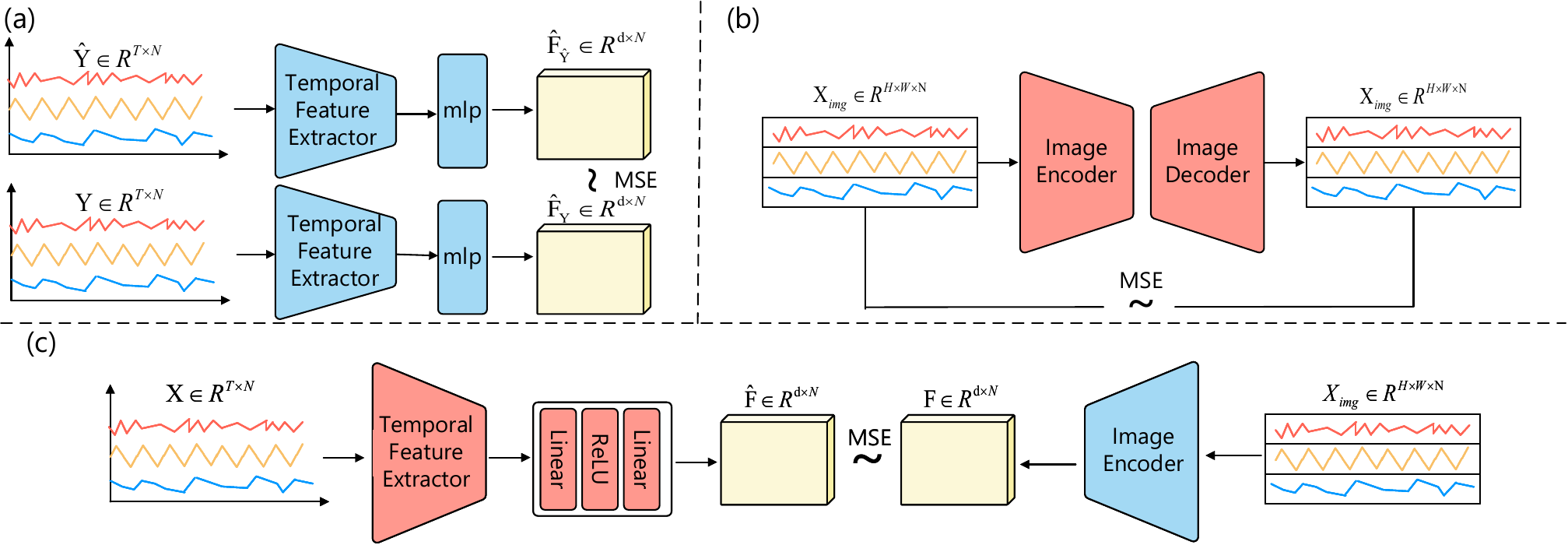}  
    \caption{Overall process of computing the geometric perception loss function. Yellow indicates features, gray represents model architecture, blue denotes frozen model parameters, and red indicates actively trained models. (a) Calculation of the perception loss function. (b) Training process of the time series image autoencoder. (c) Training process of the temporal feature extractor.}  
    \label{fig:4}  
\end{figure*}

\subsection{Frequency-Domain Loss}
To effectively model time series data, it is essential to account for periodic patterns. These patterns play a fundamental role in capturing the geometric structure of sequences, particularly for data with repetitive or oscillatory behavior. To address this, we introduce a frequency-domain loss, which operates on the spectral representations of the time series and ensures that the model learns to preserve the desired periodic characteristics.

The transformation to the frequency domain is performed using the Fast Fourier Transform (FFT)~\cite{cooley1965algorithm}. Let $\mathbf{x}$ and $\mathbf{y}$ denote the predicted sequence and the ground truth, respectively. Their frequency-domain representations are given by $\text{FFT}(\mathbf{x})$ and $\text{FFT}(\mathbf{y})$. To reduce the influence of noise and focus on the dominant frequency components, we select the top-$k$ frequencies from $\text{FFT}(\mathbf{y})$, where $k$ is proportional to the sequence length. Frequency components outside the top-$k$ set are treated as noise and are penalized in the loss function to encourage their magnitudes to approach zero.

First, the term responsible for capturing the dominant frequencies is defined as:
\begin{equation}
\mathcal{L}_\text{dom} = \sum_{f \in \mathcal{F}_\text{dom}} \left| \text{FFT}(\mathbf{x})_f - \text{FFT}(\mathbf{y})_f \right|.
\end{equation}

Next, the term for suppressing non-dominant frequency components is:
\begin{equation}
\mathcal{L}_\text{noise} = \sum_{f \notin \mathcal{F}_\text{dom}} \left| \text{FFT}(\mathbf{x})_f \right|.
\end{equation}

Finally, the complete frequency-domain loss is expressed as:
\begin{equation}
\mathcal{L}_\text{freq} = \frac{1}{\sqrt{T}} \left( \mathcal{L}_\text{dom} + \mathcal{L}_\text{noise} \right),
\end{equation}
where $\mathcal{F}_\text{dom}$ represents the set of top-$k$ dominant frequencies chosen from the ground truth spectrum. The first term ensures that the model accurately captures the dominant periodic components, while the second term suppresses irrelevant noise by minimizing non-dominant frequency magnitudes.

\subsection{Perceptual Feature Loss}

The third component addresses the challenge that temporal data, consisting of only a single time dimension, lack the rich geometric structure present in image data. This geometric structure is crucial for capturing complex relationships within the data. To overcome this limitation, we design a perceptual loss that allows the model to learn representations consistent with the geometric features of the data. Specifically, we train a feature extractor to extract such structural features and use it to compute the loss as the MSE between the features of the predicted output and the ground truth. The feature extractor is constructed in two stages: \textbf{Stage One} focuses on learning geometric features of time series image data by training an autoencoder. \textbf{Stage Two} aligns the temporal features of the time series with the geometric features learned from the image representations, enabling the computation of structural differences directly on temporal features.

The application of the perceptual loss function, along with the training process and structure, is illustrated in Figure~\ref{fig:4}.
\subsubsection{Stage One: Learning Geometric Features from Time Series Images}
\hspace*{\fill}
\\
In the first stage, we construct a temporal image autoencoder \cite{michelucci2022introduction} to learn geometric structural features from time series image data. Since there is a significant semantic gap between natural images and time series images, we train the autoencoder specifically on time series images instead of using pre-trained image models. The autoencoder is composed of an encoder that extracts latent features from the input image and a decoder that reconstructs the original image from these features.

Given a time series $\mathbf{x} \in \mathbb{R}^{T \times N}$, where $T$ is the sequence length and $N$ is the number of variables, the temporal sequence is first transformed into a corresponding image representation, denoted $\text{img}(\mathbf{x}) \in \mathbb{R}^{H \times W \times N}$, where $H$, $W$, and $N$ are the height, width, and number of variables in the generated image. The encoder is a four-layer convolutional neural network that extracts latent features from the input image., maps the input image $\text{img}(\mathbf{x})$ into a latent vector $\mathbf{z} \in \mathbb{R}^{d_z}$: $\mathbf{z} = f_\text{enc}(\text{img}(\mathbf{x})),$ where $f_\text{enc}$ represents the encoder. The decoder, consisting of a fully connected layer followed by a transposed CNN, reconstructs the input image from the latent vector: $\hat{\text{img}}(\mathbf{x}) = f_\text{dec}(\mathbf{z}),$ where $f_\text{dec}$ is the decoder network. The autoencoder is trained using the MSE between the input and reconstructed image:
\begin{equation}
\mathcal{L}_\text{ae} = \|\text{img}(\mathbf{x}) - \hat{\text{img}}(\mathbf{x})\|_2^2.
\end{equation}
Once the autoencoder is trained, the encoder $f_\text{enc}$ is frozen and used in the second stage as a feature extractor for image-like representations.

\subsubsection{Stage Two: Aligning Temporal Features with Image Features}
\hspace*{\fill}
\\
In the second stage, we train a temporal feature extractor to align the features of time series data with the geometric features learned from the autoencoder. This alignment ensures that the temporal features capture the same structural information as the image features, enabling the computation of structural differences directly on temporal data.

The temporal feature extractor consists of two components: (1) a Transformer block that captures temporal dependencies in the sequence, and (2) a two-layer MLP that maps the Transformer output into a feature vector $\mathbf{z}_\text{time} \in \mathbb{R}^{d_z \times N}$. For a given time series $\mathbf{x}$, the temporal features are extracted as:
\begin{equation}
\mathbf{z}_\text{time} = f_\text{time}(\mathbf{x}),
\end{equation}
where $f_\text{time}$ denotes the temporal feature extractor. The training objective is to minimize the difference between the temporal feature vector $\mathbf{z}_\text{time}$ and the image feature vector extracted by the encoder $f_\text{enc}$ from the corresponding time series image:
\begin{equation}
\mathcal{L}_\text{time} = \|\mathbf{z}_\text{time} - f_\text{enc}(\text{img}(\mathbf{x}))\|_2^2.
\end{equation}
Once trained, the temporal feature extractor $f_\text{time}$ is frozen and used in the perceptual loss computation. After alignment, the temporal features contain the geometric structural information learned from the time series images. This allows us to compute geometric structural differences directly on temporal features without converting the time series back into images.

\subsubsection{Perceptual Feature Loss}
\hspace*{\fill}
\\
Once the temporal feature extractor is trained and frozen, the perceptual feature loss is defined. During model training, the predicted sequence $\mathbf{x}$ and the ground truth sequence $\mathbf{y}$ are passed through the temporal feature extractor $f_\text{time}$. The extracted features are compared to enforce consistency in their learned geometric representations. Let $\mathbf{z}_x = f_\text{time}(\mathbf{x})$ and $\mathbf{z}_y = f_\text{time}(\mathbf{y})$ represent the extracted features of the predicted and ground truth sequences, respectively. The perceptual feature loss is then defined as:
\begin{equation}
\mathcal{L}_\text{perceptual} = \frac{1}{d_z} \|\mathbf{z}_x - \mathbf{z}_y\|_2^2.
\end{equation}
This loss ensures that the predicted time series captures the same structural information as the ground truth.

\subsection{Overall Loss Function}
The SATL consists of three geometric components to preserve structural patterns in time series forecasting:

\begin{equation}
\mathcal{L}_{\text{SATL}}(x,y) = \alpha \mathcal{L}_{\text{diff}}(x,y) + \beta \mathcal{L}_{\text{freq}}(x,y) + \gamma \mathcal{L}_{\text{perceptual}}(x,y)
\end{equation}
where the three terms maintain different geometric structures as previously defined, $x$ represents the predicted sequence, and $y$ denotes the ground truth sequence. To ensure both structural preservation and numerical accuracy, we combine SATL with MSE to form the overall loss function:

\begin{equation}
\mathcal{L}_{\text{total}}(x,y) = \mathcal{L}_{\text{SATL}}(x,y) + \delta \mathcal{L}_{\text{MSE}}(x,y)
\end{equation}
The hyperparameters ($\alpha$, $\beta$, $\gamma$, $\delta$) balance these competing objectives.

\section{Experiments}
\begin{table*}[t]
    \centering
    \caption{Average results for long-term forecasting tasks with prediction lengths {24,36,48,60} for ILI and \{96,192,336,720\} for others. Metrics include MSE, MAE, and TGSI, averaged across prediction lengths. Input sequence lengths are $T=96$ except $T=36$ for ILI. Avg Improved measures SATL's average improvements over MSE training. The better metrics are colored in \textcolor{red}{red}.}
    \resizebox{\textwidth}{!}{
    \setlength{\tabcolsep}{1pt}
    \begin{tabular}{c|ccc|ccc||ccc|ccc||ccc|ccc||ccc|ccc||ccc}
    \toprule
    \multicolumn{1}{c|}{Model} & \multicolumn{6}{c||}{TimeMixer(\cite{wang2024timemixer})} & \multicolumn{6}{c||}{PatchTST(\cite{nie2022time})} & \multicolumn{6}{c||}{TimesNet(\cite{wu2022timesnet})}  & \multicolumn{6}{c||}{Autoformer(\cite{wu2021autoformer})} & \multicolumn{3}{c}{ -} \\
    \midrule
    \multicolumn{1}{c|}{Loss} & \multicolumn{3}{c|}{MSE} & \multicolumn{3}{c||}{+SATL} & \multicolumn{3}{c|}{MSE} & \multicolumn{3}{c||}{+SATL} & \multicolumn{3}{c|}{MSE} & \multicolumn{3}{c||}{+SATL} & \multicolumn{3}{c|}{MSE} & \multicolumn{3}{c||}{+SATL}  & \multicolumn{3}{c}{ Avg Improved } \\
    \midrule
     \multicolumn{1}{c|}{Metric}                                        & MSE   & MAE  & TGSI & MSE   & MAE  & TGSI & MSE   & MAE  & TGSI & MSE   & MAE  & TGSI & MSE   & MAE  & TGSI  & MSE   & MAE  & TGSI & MSE   & MAE  & TGSI & MSE   & MAE  & TGSI & MSE   & MAE  & TGSI \\ 
    \midrule[\heavyrulewidth]
    
    \multirow{1}{*}{{ETTh1}}    & 0.448 & 0.440 & 0.6125 & \textcolor{red}{0.433} & \textcolor{red}{0.431} & \textcolor{red}{0.6184} & 0.449 & 0.448 & 0.6010 & \textcolor{red}{0.436}& \textcolor{red}{0.437} & \textcolor{red}{0.6075} & 0.460 & 0.454 & 0.5816 & \textcolor{red}{0.449} & \textcolor{red}{0.447} & \textcolor{red}{0.5903} & 0.519 & 0.499 & 0.5234 & \textcolor{red}{0.472} & \textcolor{red}{0.471} & \textcolor{red}{0.5508} & 4.42\% & 2.91\% & 2.19\% \\
    \midrule
    
    \multirow{1}{*}{{ETTh2}}    & 0.391 & 0.411 & 0.5814 & \textcolor{red}{0.370} & \textcolor{red}{0.394} & \textcolor{red}{0.5979} & 0.384 & 0.411 & 0.5790 & \textcolor{red}{0.367} & \textcolor{red}{0.395} & \textcolor{red}{0.5945} & 0.406 & 0.420 & 0.5522 & \textcolor{red}{0.382} & \textcolor{red}{0.404} & \textcolor{red}{0.5923} & 0.451 & 0.462 & 0.5201 & \textcolor{red}{0.428} & \textcolor{red}{0.445} & \textcolor{red}{0.5372} & 5.20\% & 3.88\% & 4.02\% \\    
    \midrule
    
    \multirow{1}{*}{{ETTm1}}    & 0.381 & 0.397 & 0.5800 & \textcolor{red}{0.380} & \textcolor{red}{0.394} & \textcolor{red}{0.5808} & 0.389 & 0.403 & 0.5895 & \textcolor{red}{0.385} & \textcolor{red}{0.398} & \textcolor{red}{0.5918} & 0.410 & 0.418 & 0.5601 & \textcolor{red}{0.404} & \textcolor{red}{0.413} & \textcolor{red}{0.5638} & 0.615 & 0.527 & 0.4552 & \textcolor{red}{0.517} & \textcolor{red}{0.489} & \textcolor{red}{0.4776} & 4.67\% & 2.60\% & 1.53\% \\
    \midrule
    
    \multirow{1}{*}{{ETTm2}}    & 0.276 & {0.322} & 0.6154 & \textcolor{red}{0.274} & \textcolor{red}{0.321} & \textcolor{red}{0.6177} & 0.291 & 0.334 & 0.6005 & \textcolor{red}{0.284} & \textcolor{red}{0.328} & \textcolor{red}{0.6107} & 0.299 & 0.333 & 0.6064 & \textcolor{red}{0.292} & \textcolor{red}{0.328} & \textcolor{red}{0.6109} & 0.327 & 0.364 & 0.5637 & \textcolor{red}{0.317} & \textcolor{red}{0.359} & \textcolor{red}{0.5705} & 2.13\% & 1.25\% & 1.01\% \\    
    \midrule

    \multirow{1}{*}{{Exchange}} & 0.375 & 0.410 & 0.6383 & \textcolor{red}{0.366} & \textcolor{red}{0.404} & \textcolor{red}{0.6389} & 0.388 & 0.416 & 0.6380 & \textcolor{red}{0.363} & \textcolor{red}{0.406} & \textcolor{red}{0.6387} & 0.410 & 0.442 & 0.6224 & \textcolor{red}{0.403} & \textcolor{red}{0.434} & \textcolor{red}{0.6243} & 0.507 & 0.501 & 0.6006 & \textcolor{red}{0.494} & \textcolor{red}{0.492} & \textcolor{red}{0.6030} & 3.28\% & 1.87\% & 0.23\% \\
    \midrule

    \multirow{1}{*}{{Weather}} & 0.243 & 0.274 & 0.5892 & \textcolor{red}{0.242} & \textcolor{red}{0.272} & \textcolor{red}{0.5909} & 0.257 & 0.278 & 0.5817 & \textcolor{red}{0.255} & \textcolor{red}{0.277} & \textcolor{red}{0.5821} & 0.258 & 0.285 & 0.5686 & \textcolor{red}{0.256} & \textcolor{red}{0.284} & \textcolor{red}{0.5710} & 0.349 & 0.383 & 0.4125 & \textcolor{red}{0.332} & \textcolor{red}{0.370} & \textcolor{red}{0.4334} & 1.71\% & 1.21\% & 1.46\%\\
    \midrule

    \multirow{1}{*}{{Electricity}} & 0.185 & 0.274 & 0.8118 & \textcolor{red}{0.184} & \textcolor{red}{0.273} & \textcolor{red}{0.8129} & 0.202 & 0.293 & 0.8021 & \textcolor{red}{0.197}& \textcolor{red}{0.286}& \textcolor{red}{0.8061} & 0.192& 0.294 & 0.7970 & \textcolor{red}{0.191}& \textcolor{red}{0.292}& \textcolor{red}{0.7985} & 0.235 & 0.345 & 0.7540 & \textcolor{red}{0.212}& \textcolor{red}{0.323}& \textcolor{red}{0.7728} & 3.33\% & 2.45\% & 0.83\% \\
    \midrule

    \multirow{1}{*}{{ILI}}& 2.323 & 0.956                & 0.7157                & \textcolor{red}{2.294} & \textcolor{red}{0.940} & \textcolor{red}{0.7192} & 2.145                & 0.897                & 0.7300                & \textcolor{red}{2.127} & \textcolor{red}{0.893} & \textcolor{red}{0.7305} & 2.262 & 0.928                & 0.7180                & \textcolor{red}{2.095} & \textcolor{red}{0.910} & \textcolor{red}{0.7197} & 3.129                & {1.205} & 0.6316 & \textcolor{red}{3.086} & \textcolor{red}{1.198} & \textcolor{red}{0.6386} & 2.71\% & 1.16\% & 0.48\% \\
    \midrule
    
    \multirow{1}{*}{{Avg Improved}}& - & -                & -                & 1.79\% & 1.43\% & 0.66\% & -                & -                & -               & 2.66\% & 1.87\% & 0.82\% & - & - & -                & 2.81\% & 3.61\% & 1.41\% & -                & - & - & 6.47\% & 3.75\% & 2.96\% & & & \\

    \bottomrule
    \end{tabular}}
    \label{tab:1}
\end{table*}

\begin{figure*}[htbp]  
    \centering      
    \includegraphics[width=0.99\textwidth]{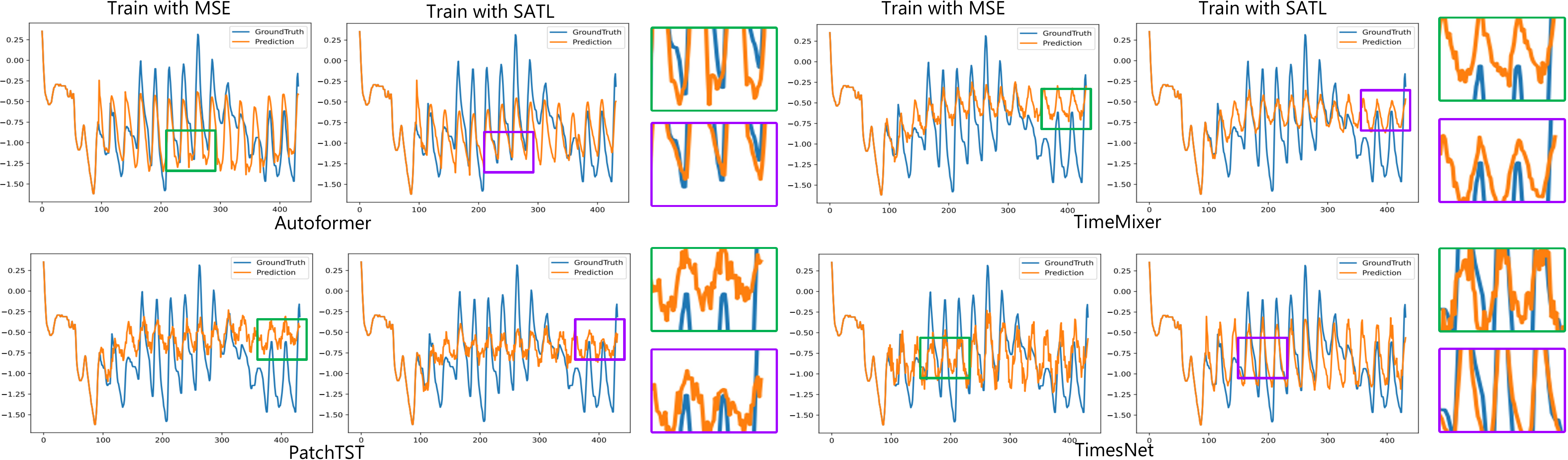}  
    \caption{Visualization of time series forecasting tasks on the ETTh2 dataset for the 96\_336 task. Each column is divided into two sections: the left side displays results from models trained with MSE, while the right side shows outcomes from models trained with SATL. The enlarged sections highlight that models trained with SATL exhibit significantly enhanced geometric structure similarity in their predictions.}  
    \label{fig:5}  
\end{figure*}

\subsection{Experimental Setup}

\textbf{Baselines}. To evaluate the effectiveness of the proposed SATL in time series forecasting, we select 4 baseline models: TimeMixer\cite{wang2024timemixer}, PatchTST\cite{nie2022time}, TimesNet\cite{wu2022timesnet}, and Autoformer\cite{wu2021autoformer}. These models represent a range of architectures and performance levels, enabling us to assess the generalizability of SATL across different approaches to time series forecasting.

\textbf{Datasets}. We conduct experiments on eight publicly available time series datasets: ETTh1, ETTh2, ETTm1, ETTm2\cite{haoyietal-informer-2021}, Weather, Exchange\cite{wu2021autoformer}, Electricity\cite{lai2018modeling} and ILI. These datasets are widely used benchmarks in time series forecasting, covering diverse application scenarios and temporal patterns. For each dataset, we follow the standard data preprocessing pipeline, including normalization and train-validation-test splits, to ensure consistency.

\textbf{Metrics}. We adopt three evaluation metrics: Mean Squared Error (MSE) and Mean Absolute Error (MAE) as traditional numerical metrics, and our proposed TGSI as a geometric structure evaluation metric. This combination ensures a comprehensive assessment of both numerical accuracy and geometric pattern preservation.

\textbf{Settings}. To ensure fair comparisons, all experiments are conducted under the same settings, with all hyperparameters fixed between training runs, except for the loss function. We maintain the same initialization, learning rate, batch size, and other training parameters. For the majority of datasets, we fix the hyperparameters of the loss functions at $\alpha = 0.2$, $\beta = 0.2$, $\gamma = 0.1$, and $\delta = 0.5$. For the remaining datasets, these hyperparameters are selected based on their respective validation sets.  Additionally, to produce TGSI results for models trained with MSE and fill in missing baseline results on certain datasets, \textbf{we re-train all baselines ourselves}. 

For the transformation of time series to image, we set the image width equal to the time series length with a fixed height of 200 pixels, where each time point is vertically extended by $d=100$ pixels. In SATL training with Perceptual Feature Loss, we conduct 30 epochs for stage one and 10 epochs for stage two, using Adam optimizer with a learning rate of 0.001 and batch size of 16 throughout both stages. All models are implemented using PyTorch and trained on a machine equipped with an NVIDIA RTX4090 GPU.

\subsection{Comparative Performance Results}
As shown in Table~\ref{tab:1}, the quantitative comparison reveals that models trained with SATL consistently achieve superior performance compared to MSE-trianed models across all datasets and metrics. Importantly, this improvement comes \textbf{without any additional computational cost during inference}. The results demonstrate SATL's effectiveness in enhancing both numerical accuracy and geometric structure preservation in time series forecasting. Notably, SATL achieves 5.20\% MSE improvement on the ETTh2 dataset and 6.42\% performance gain in Autoformer. Importantly, SATL enables models to explore capabilities beyond their original designs - while PatchTST initially underperformed TimeMixer on the Exchange dataset, our approach allowed it to achieve superior results.

We visualized the prediction results for the ETTh2 dataset in Figure~\ref{fig:5}, focusing on the task of forecasting 336 steps with an input sequence length of 96. The left column displays the results obtained using MSE for training, while the right column presents the results from training with SATL. The enlarged details on the right reveal that the predictions made by the model trained with SATL exhibit a geometric structure that is more closely aligned with the ground truth.

\subsection{Comparison with Other Loss Functions}
\begin{figure}[h]  
    \centering      
    \includegraphics[width=0.49\textwidth]{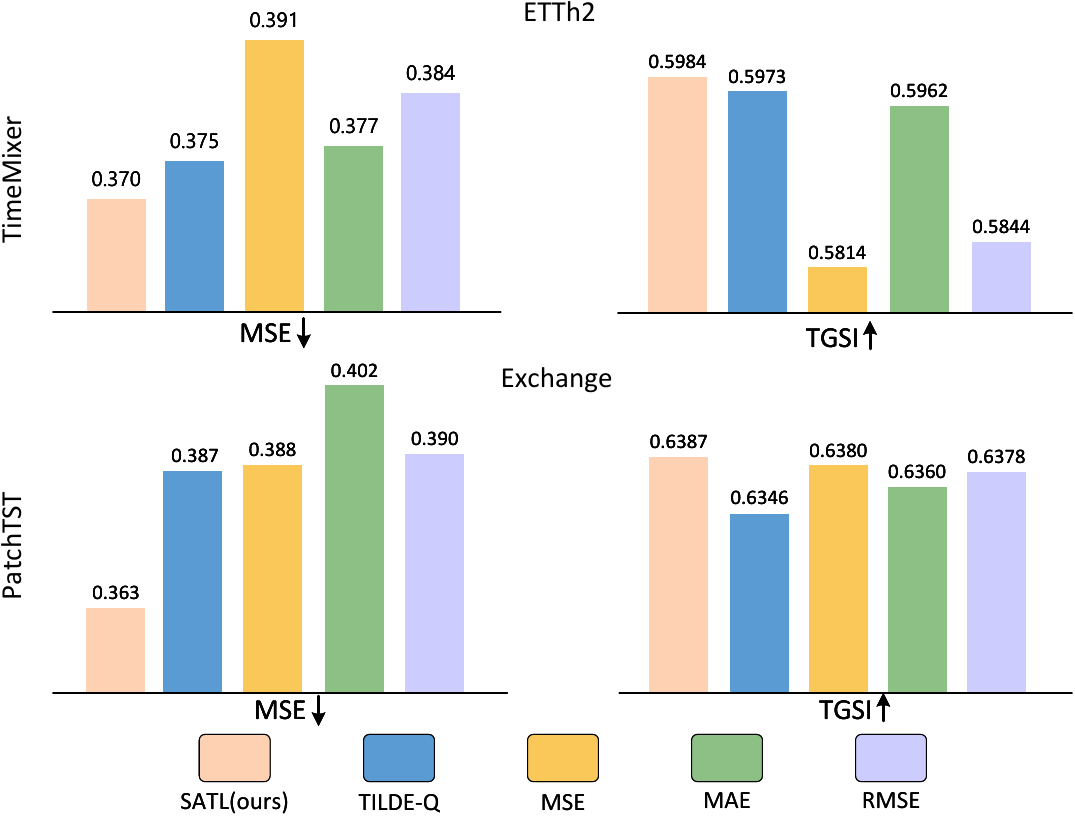}  
    \caption{Comparison of different loss functions}  
    \label{fig:7}  
    \vspace{-5mm}
\end{figure}

As illustrated in Figure~\ref{fig:7}, we compare SATL against various baseline loss functions: MSE, MAE, Root Mean Squared Error(RMSE) and TILDE-Q. The controlled experiments employ TimeMixer on ETTh2 and PatchTST on Exchange datasets, maintaining identical configurations except for the loss functions. SATL achieves the lowest MSE values and highest TGSI scores among all compared methods, demonstrating its dual advantage in both numerical accuracy and geometric structure preservation. 

\subsection{Ablation Study}

\begin{figure*}[t]
\centering 
\includegraphics[width=0.95\textwidth]{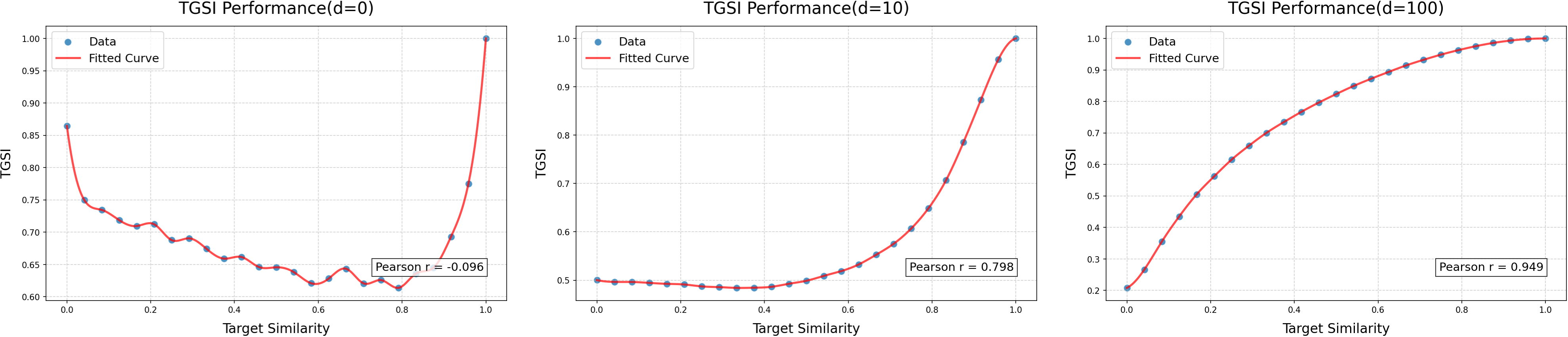} 
    \caption{Fitted curve of TGSI versus sample similarity for varying vertical extension lengths $d$}  
    \label{fig:6}  
\end{figure*}

\begin{table*}[t]
    \centering
    \caption{Ablation study of SATL components: Difference (First-Order Temporal Loss), Frequency (Frequency-Domain Loss), Perceptual (Perceptual Feature Loss), and MSE (Mean Squared Error component)}
    \resizebox{\textwidth}{!}{
        \begin{tabular}{c|c|c|c|c|c|c|c|c}
            \hline
             Difference & Frequency & Perceptual & MSE  & 96 & 192 & 336 & 720 & Avg. \\
            \midrule
            \checkmark & \checkmark & \checkmark & \checkmark & \textbf{0.379}/0.6336 & \textbf{0.424}/0.6042 & \textbf{0.469}/\textbf{0.5944} & \textbf{0.472}/\textbf{0.5977} & \textbf{0.436}/\textbf{0.6075} \\
            \midrule
            \checkmark & \checkmark & \checkmark & $\times$ & 0.383/0.6242 & 0.428/\textbf{0.6144} & 0.515/0.5695 & 0.578/0.5394 & 0.476/0.5859 \\
            \midrule
            \checkmark & \checkmark & $\times$ & \checkmark & 0.380/\textbf{0.6344} & 0.431/0.6013 & 0.478/0.5880 & 0.497/0.5777 & 0.447/0.6003 \\
            \midrule
            \checkmark & $\times$ & \checkmark & \checkmark & 0.388/0.6328 & 0.430/0.6036 & 0.557/0.5338 & 0.566/0.5369 & 0.485/0.5768 \\
            \midrule
            $\times$ & \checkmark & \checkmark & \checkmark & 0.381/0.6333 & 0.428/0.6029 & 0.506/0.5790 & 0.590/0.5363 & 0.476/0.5879 \\
            \midrule
            $\times$ & $\times$ & $\times$ & \checkmark & \textbf{0.379}/0.6324 & 0.425/0.6025 & 0.470/0.5906 & 0.521/0.5783 & 0.449/0.6010 \\
            \midrule
        
        \end{tabular}
        }
        \label{tab:2}
\end{table*}

To systematically investigate the impact of individual components in SATL and their sensitivity to different forecasting horizons, we conduct comprehensive ablation experiments on the ETTh1 dataset using PatchTST architecture, while keeping all other experimental settings identical. As presented in Table~\ref{tab:2}, the performance degrades when either component is removed, with particularly significant deterioration observed upon eliminating the MSE term. This substantiates the essential role of maintaining fundamental numerical similarity in the loss function. 

Furthermore, the experiments reveal distinct behavioral patterns across varying prediction lengths: while differences among variants remain marginal for shorter-term forecasting (96 steps), the performance gap becomes substantially more pronounced for long-term predictions (720 steps). This demonstrates increasing sensitivity to loss function design as the forecasting horizon extends, highlighting SATL's particular advantage in long-range time series forecasting scenarios.

\subsection{Validation of TGSI Metric}

To systematically validate the effectiveness of TGSI and demonstrate the necessity of vertical expansion in the time-to-image transformation, we designed a controlled experimental framework. As shown in Figure~\ref{fig:6}, our evaluation begins with generating a base reference sequence $y$ composed of multiple periodic components with strategically injected noise to simulate real-world temporal patterns. 

We then generate test sequences $\{x_i\}$ with precisely controlled similarity levels $p \in [0,1]$ through three distinct deformation operators: amplitude scaling ($\mathcal{T}_1(x) = p \cdot y$), constant offset ($\mathcal{T}_2(x) = y + (1-p)\cdot c$), and adaptive noise injection ($\mathcal{T}_3(x) = y + \mathcal{N}(0,1-p)$). For each similarity level $p$, the final TGSI score is computed as the mean value across all three deformation types: $\mathrm{TGSI}(p) = \frac{1}{3}\sum_{i=1}^3 \mathrm{TGSI}(y, \mathcal{T}_i(x))$.

The evaluation tests three vertical expansion configurations ($d=0$, $d=10$, and $d=100$) to analyze the structural encoding capability. The results reveal three critical observations: First, without vertical expansion ($d=0$), TGSI fails to establish meaningful structural relationships, exhibiting non-monotonic behavior with a Pearson correlation of only $r=-0.1$. Second, moderate expansion ($d=10$) shows limited discriminative power, particularly in low-similarity regions ($p<0.6$), where TGSI scores completely lose their ability to differentiate between sequences. Most significantly, full expansion ($d=100$) achieves near-perfect monotonicity ($r=0.95$) with clear differentiation across all similarity levels, confirming that adequate vertical expansion is essential for capturing temporal geometric structures.

These findings highlight the importance of sufficient vertical expansion width $d$, which provides the spatial context necessary to capture long-range temporal dependencies while preserving structural features. Larger values of $d$ improve the robustness and discriminative power of TGSI, whereas shorter expansions fail to distinguish between structural similarities and random variations.

\section{Conclusion}
In this paper, we present TGSI, a novel geometric structure evaluation metric for time series forecasting, along with SATL, a shape-aware training loss function that effectively incorporates structural information into model optimization. Our approach overcomes the limitations of conventional metrics such as MSE by establishing a comprehensive framework that simultaneously considers numerical accuracy and geometric fidelity. Extensive experimental results demonstrate that SATL significantly enhances model performance across both traditional numerical metrics and our proposed geometric evaluations, while maintaining computational efficiency during inference. This work makes contributions by highlighting the role of geometric structure in time series analysis, thereby opening up new research directions in structure-aware time series modeling.

\bibliographystyle{ACM-Reference-Format}
\bibliography{main}

\end{document}